%% file: main.tex
\documentclass[10pt,twocolumn,letterpaper]{article}

\usepackage{cvpr}              % To produce the CAMERA-READY version

\usepackage{CJK}
\usepackage{multicol}
\usepackage{color}
\usepackage{xcolor}
\usepackage{colortbl}
\usepackage{bm}
\usepackage{extarrows}
\usepackage{float}
\usepackage{bbm}
\usepackage{booktabs}
\usepackage{multirow}
\usepackage{amsmath}
\usepackage{amssymb}
\usepackage{algorithm}
\usepackage{algorithmic}
\usepackage{makecell}
\usepackage{amsmath}
\usepackage{amssymb}
\usepackage{amsthm}
\usepackage[utf8]{inputenc}
\usepackage{amsfonts} % \mathbf (z) 符号
\usepackage{caption} 
\usepackage{newfloat}
\usepackage{listings}

\usepackage{xcolor}
\definecolor{thmcolor}{HTML}{0201F5}

\newtheoremstyle{coloredthm}% name
  {3pt}%      Space above
  {3pt}%      Space below
  {\itshape}%         Body font
  {}%         Indent amount (empty = no indent, \parindent = para indent)
  {\bfseries\color{thmcolor}}% Thm head font (Bold AND Colored)
  {.}%        Punctuation after thm head
  {.5em}%     Space after thm head: " " = normal interword space;
  {}%         Thm head spec (can be left empty, meaning 'normal')

\theoremstyle{coloredthm}

\definecolor{cvprblue}{rgb}{0.21,0.49,0.74}
\definecolor{supplcolor}{HTML}{A32D26}

\definecolor{noise20color}{HTML}{FFF9E6} % 极淡橙
\definecolor{noise50color}{HTML}{EDF7ED} % 极淡绿
\definecolor{noise80color}{HTML}{EBF5FB} % 极淡蓝

\usepackage[pagebackref,breaklinks,colorlinks,citecolor=brown]{hyperref}
\usepackage{thmtools}

\def\paperID{*****} % *** Enter the Paper ID here
\def\confName{CVPR}
\def\confYear{2026}

\newcommand{\method}{OmniEgo-R$^2$}
\newcommand{\methodfull}{\textbf{Omni}domain \textbf{Ego}centric \textbf{R}outed \textbf{R}easoning}

\title{OmniEgo-R$^2$: A Routed Reasoning Framework for the 1st Cross-Domain EgoCross Challenge at CVPR 2026}

\author{Zixu Li$^{1}$~~Zhiwei Chen$^1$~~Zhiheng Fu$^1$~~Wenbo Wang$^{1}$~~Yupeng Hu$^{1}$~~Weili Guan$^{2}$~~Liqiang Nie$^2$ \vspace{2mm}\\
$^1$Shandong University\hspace{1.5cm}$^2$Harbin Institute of Technology (Shenzhen)\hspace{1.5cm}\\
{\tt\small \{lizixu.cs,zivczw,fuzhiheng8,honeyguan,nieliqiang\}@gmail.com;} \\ 
{\tt\small wangwenbo@mail.sdu.edu.cn,huyupeng@sdu.edu.cn}\\
}

\begin{document}
% \begin{CJK}{UTF8}{gbsn}
\maketitle
\input{2-sec/0_abstract}

\input{2-sec/1_introduction}

\input{2-sec/2_method}

\input{2-sec/3_exp}

\input{2-sec/4_conclusion}

% \clearpage
{
    \small
    \bibliographystyle{ieeenat_fullname}
    \bibliography{main}
}

% WARNING: do not forget to delete the supplementary pages from your submission 
% \end{CJK}
\end{document}

%% file: 2-sec/0_abstract.tex
\begin{abstract}
The 1st Cross-Domain EgoCross Challenge at EgoVis, CVPR 2026 evaluates whether multimodal large language models can reason over egocentric videos across surgery, industry, extreme sports, and animal perspective. We achieved second place in both the Source-Limited and Open-Source tracks. In this report, we formulate EgoCross as a robust cross-domain embodied video reasoning problem rather than a simple multiple-choice visual question answering task. We identify three key challenges: (C1) temporal boundary ambiguity, where critical state transitions are sparsely sampled and often occur between frames; (C2) cross-domain semantic granularity mismatch, where the same capability requires different domain-specific visual grammar; and (C3) decision instability under close options, where long multimodal reasoning can select unsupported distractors or produce malformed outputs. To address them, we propose \method{} (\methodfull), a unified routed reasoning pipeline consisting of temporal-evidence normalization, domain-agnostic capability routing, structured perception--dynamics--decision reasoning, boundary-aware option verification, and defensive answer calibration. \method{} uses the Qwen3-VL-4B-SFT checkpoints on each EgoCross domain as the visual-language backbone, and wraps them with lightweight test-time reasoning and parsing programs. Our final submissions obtain 66.35\% overall accuracy in the Source-Limited track and 66.77\% in the Open-Source track, ranking second in both leaderboards. The codes are available on \href{https://github.com/Lee-zixu/OmniEgo-R2}{https://github.com/Lee-zixu/OmniEgo-R2}
\end{abstract}

%% file: 2-sec/1_introduction.tex
\section{Introduction}
With the development of multimodal learning~\cite{qwen3technicalreport,OFFSET,PAIR,ENCODER}, egocentric video understanding has been studied through dedicated benchmarks and models~\cite{egovqa,egoschema,egovlpv2,EgoAdapt}, but it remains fundamentally different from conventional third-person video recognition~\cite{Air-Know,ConeSep,MEDIAN,ERASE}. In first-person videos, the camera is attached to the acting subject, so the decisive evidence is often partial, unstable, and action-dependent~\cite{egogpt,egothink,egotextvqa}: an object may appear only at the image boundary, a phase change may be encoded by a subtle motion transition, and a target may even be absent from most sampled frames. EgoCross~\cite{egocross} makes this setting more demanding by spanning four domains---surgery, industry, XSports, and animal perspective---and 15 sub-task types that stress different capabilities. The task distribution is highly heterogeneous: surgery emphasizes dominant held-object identification, object spatial localization, object counting, object not visible identification, and temporal localization; industry concentrates on object counting, next interaction prediction, held-object identification, not-visible reasoning, and localization; animal perspective is dominated by interaction identification and interaction temporal localization; and XSports focuses on action temporal localization, next direction prediction, action sequence identification, special action identification, and sport identification. This composition shows that EgoCross is not a single capability benchmark but a mixture of fine-grained perception, temporal boundary reasoning, and cross-domain semantic grounding. Consequently, advancing research on such a comprehensive benchmark holds great potential to facilitate downstream applications in related fields, including composition reasoning~\cite{egotextvqa,FineCIR,qwen25vl,HINT,qwenvl,internvl,TEMA,MELT}, video understanding~\cite{ReTrack,egothink,HUD,REFINE}, and multimodal learning~\cite{STABLE,gpt4,INTENT,HABIT,TempRet}.

Direct end-to-end prompting on such data exposes three challenges. \textbf{C1: Temporal boundary ambiguity.} Many EgoCross questions ask \emph{when} a phase, interaction, or motion pattern starts, but the sampled frames are sparse and the decisive transition may happen between frames. This is especially common in surgery and XSports, where action temporal localization and phase/direction prediction dominate. \textbf{C2: Cross-domain semantic granularity mismatch.} The same abstract capability must be instantiated with different visual grammar across domains: a ``held object'' means a surgical tool in surgery, an industrial component in ENIGMA, and a moving target in animal perspective; similarly, spatial localization in industry requires small-object inspection, while in XSports it depends on body-motion cues and horizon changes. \textbf{C3: Decision instability under close options.} Many subtasks are closed-set multiple choice with semantically similar distractors or ``not visible'' hypotheses, so models may produce plausible reasoning but still select an unsupported option or return malformed output. These failures are amplified by cross-domain shifts and by the need to solve identification, localization, prediction, and counting within one unified interface.

To address these challenges, we propose \method{} (\methodfull), a unified routed reasoning pipeline. Instead of designing dataset-specific heuristics, \method{} decomposes each sample into evidence normalization, capability routing, role-based reasoning, option-boundary verification, and answer calibration. The key idea is to treat each domain as a different semantic basis plugged into the same reasoning program, so that the model reasons over surgery tools, industrial parts, animal interactions, and sports trajectories with a common decision architecture.

Our contributions are threefold, summarized as follows:
\begin{itemize}
    \item We analyze EgoCross through its task composition and identify three concrete bottlenecks---temporal boundary ambiguity, cross-domain semantic granularity mismatch, and decision instability under close options, that explain why a generic MLLM prompt is insufficient.
    \item We propose \method{} (\methodfull), a routed reasoning pipeline with temporal evidence normalization, capability-oriented routing, role-decomposed reasoning, boundary-aware option verification, and defensive answer calibration.
    \item We achieve second place in both the Source-Limited and Open-Source tracks, and demonstrate the effectiveness of the proposed modules with the cooperation on the official EgoCross Qwen3-VL-4B-SFT backbones.
\end{itemize}

%% file: 2-sec/2_method.tex
\section{Methodology}
\begin{figure*}[t]
    \centering
    \includegraphics[width=0.95\linewidth]{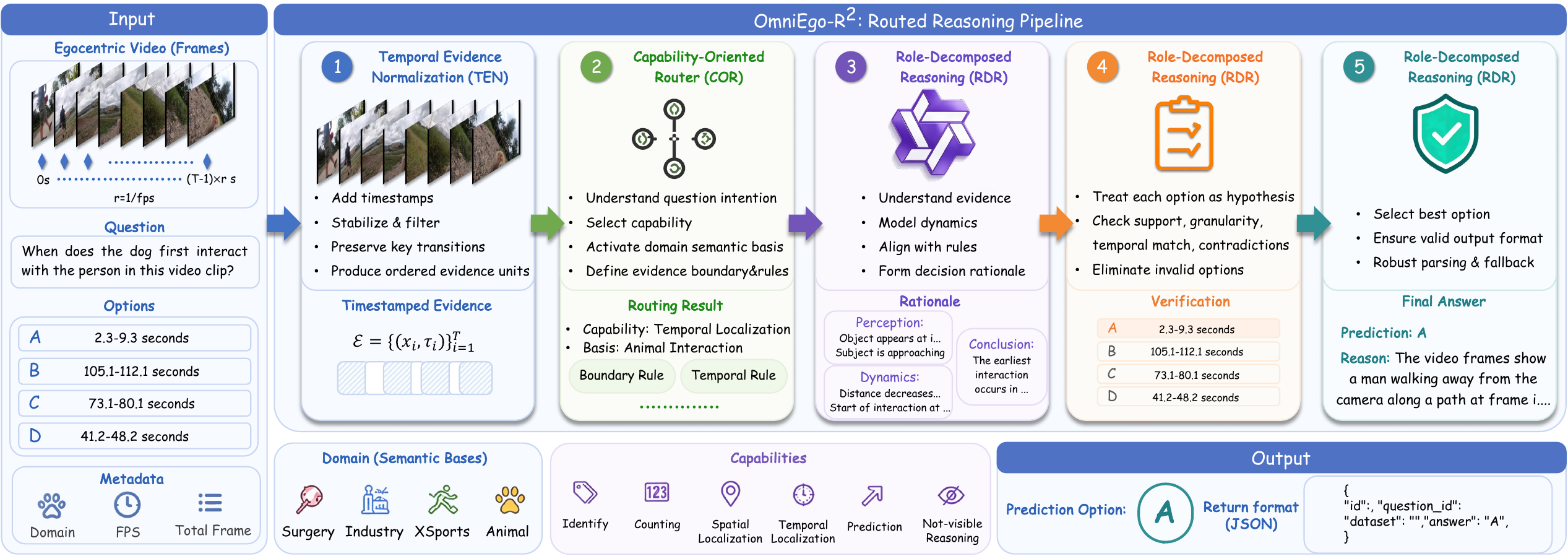}
    \vspace{-8pt}
    \caption{Overview of \method. Domains are expressed as semantic bases plugged into a shared evidence normalization, capability grounding, structured reasoning, option verification, and answer calibration pipeline.}
        \vspace{-12pt}
    \label{fig:pipeline}
\end{figure*}

\method{} (\methodfull) is a unified test-time reasoning framework built on the official domain-SFT Qwen3-VL-4B backbones~\cite{egocross,qwen3technicalreport}. The core design is to transform a heterogeneous EgoCross sample into a common evidence--capability--verification program, rather than treating the four domains as isolated scripts. As shown in Fig.~\ref{fig:pipeline}, \method{} consists of Temporal Evidence Normalization (TEN), Capability-Oriented Router (COR), Role-Decomposed Reasoning (RDR), Boundary-aware Option Verification (BOV), and Defensive Answer Calibration (DAC). Each module is instantiated with structured prompts, timestamped visual inputs, and robust output parsing. In the following, we first provide the problem formulation and them elaborate on each module of \method.

\subsection{Problem Formulation and Overview}
For each sample, the input is a frame sequence $X=\{x_i\}_{i=1}^{T}$, a question $q$, options $O=\{o_A,o_B,o_C,o_D\}$, and metadata $m$ such as domain and sampling rate. The goal is to predict $\hat{y}\in\{A,B,C,D\}$. A direct MLLM baseline estimates $p_{\theta}(y\mid X,q,O)$ from a single multimodal query, which entangles evidence localization, temporal alignment, domain semantics, option comparison, and output validity. \method{} instead decomposes prediction as
\begin{equation}
    \hat{y}=\mathcal{C}\circ\mathcal{V}\circ\mathcal{R}\circ\mathcal{G}\circ\mathcal{N}(X,q,O,m),
    \label{eq:pipeline-composition}
\end{equation}
where $\mathcal{N}$ normalizes temporal evidence, $\mathcal{G}$ grounds the question into a capability and domain semantic basis, $\mathcal{R}$ performs structured reasoning, $\mathcal{V}$ verifies candidate options, and $\mathcal{C}$ converts the verified decision into a valid option label. Notably, Eq.~\eqref{eq:pipeline-composition} is a test-time reasoning program, not a differentiable network.

\subsection{Temporal Evidence Normalization (TEN)}
EgoCross questions frequently use temporal language: an interaction may ``start'', a motion may ``change direction'', or an action may occur closest to a listed timestamp. If sampled frames are supplied only as an ordered image list, the model can confuse local frame order with global time, especially when a long sequence is processed in parts. Therefore, we design TEN to convert the frame sequence into timestamped visual units, formulated as,
% Egocentric evidence is sparse, unstable, and often temporally ambiguous. TEN first converts each sampled frame into a timestamped unit,
\begin{equation}
    \tilde{x}_i=(x_i,\tau_i),\qquad \tau_i=\frac{i+i_0}{r},
    \label{eq:timestamp}
\end{equation}
where $x_i$ is the $i$-th frame in the current segment, $\tau_i$ is its global timestamp, $i_0$ is the global offset of the segment, and $r$ is the sampling rate. This prevents later reasoning from confusing local frame order with global time, especially when long videos are processed in segments.

% TEN also imposes a reliability prior on observation. Rather than computing an explicit frame score, the model is guided to prefer frames that are both option-relevant and visually stable:
The second role of TEN is to make the model distinguish informative frames from egocentric noise. Rather than computing a numerical frame score with an external network, we impose a reliability criterion on the observation stage, formulated as,
\begin{equation}
\begin{aligned}
    x_i \succ x_j \quad \mathrm{if}\quad
    &\mathrm{Rel}(x_i,q,O) > \mathrm{Rel}(x_j,q,O),\\[-1pt]
    &\mathrm{Deg}(x_i) < \mathrm{Deg}(x_j),
\end{aligned}
\label{eq:reliability-order}
\end{equation}
where $\mathrm{Rel}$ denotes question-option relevance, $\mathrm{Deg}$ denotes blur, abrupt camera motion, or irrelevant clutter, and $x_i\succ x_j$ means that $x_i$ should be treated as stronger evidence. This is a conceptual evidence ordering induced at test time, not an explicitly computed neural score, learned weighting module, or external detector. TEN therefore outputs $\mathcal{E}=\{(x_i,\tau_i)\}_{i=1}^{T}$ together with a reliability-oriented observation rule: stable frames provide primary support, while blurred frames are retained only when they mark transitions.

\subsection{Capability-Oriented Router (COR)}
A naive way to handle cross-domain data is to prepend the dataset name to the prompt. This is insufficient because the same capability has different visual meanings across domains. Temporal localization in surgery depends on tool--tissue contact, while temporal localization in XSports may depend on trajectory change or body orientation; counting in industry requires respecting controlled grouping rules, whereas counting in surgery may depend on visible instruments or anatomical regions. Thus, we design COR to address this by decoupling the capability being tested from the domain semantic basis used to interpret evidence.

Formally, COR assigns each sample to a capability $c$ and a domain semantic basis $\mathcal{B}_d$,
\begin{equation}
    (c,\mathcal{B}_d)=\rho(q,O,m), \qquad
    \mathcal{P}=\phi(c,\mathcal{B}_d,q,O),
    \label{eq:router}
\end{equation}
where $\rho$ denotes capability-domain grounding, $d$ is the domain, and $\mathcal{P}$ is the resulting reasoning protocol. The capability $c$ covers identification, counting, spatial localization, temporal localization, prediction, and not-visible reasoning. The semantic basis $\mathcal{B}_d$ specifies the domain-specific visual grammar: tool-centric reasoning for surgery, object-centric procedural reasoning for industry, physics-centric embodied reasoning for XSports, and self-other behavioral reasoning for animal perspective.

% where $\mathcal{P}$ is the reasoning protocol. The capability space covers identification, counting, spatial localization, temporal localization, prediction, and not-visible reasoning; the semantic basis specifies how these capabilities are interpreted in each domain---tool-centric for surgery, object-centric procedural for industry, physics-centric embodied for XSports, and self-other behavioral for animal perspective.

\begin{table}[t]
    \centering
    \small
        \caption{Capability operators in \method. The evidence boundary is shared across domains, while the semantic basis determines how it is instantiated.}
        \vspace{-10pt}
        \resizebox{\linewidth}{!}{
    \begin{tabular}{p{0.27\linewidth}p{0.31\linewidth}p{0.30\linewidth}}
        \toprule
        Capability & Evidence Boundary & Verification Rule\\
        \midrule
        Identification & attributes & eliminate mismatch\\
        Counting & stable instances & obey granularity\\
        Spatial & target--anchor pair & match relation\\
        Temporal & state transition & closest timestamp\\
        Prediction & final trajectory & next-state consistency\\
        Not-visible & full-frame coverage & complete absence\\
        \bottomrule
    \end{tabular}
    }
\vspace{-13pt}
    \label{tab:operators}
\end{table}

Table~\ref{tab:operators} summarizes the reusable operators. The important design choice is that the router does not create four unrelated pipelines. It defines a shared evidence boundary for each capability and then instantiates that boundary in the correct semantic space. This provides the bridge from TEN to reasoning: once frames are timestamped and reliability-oriented, COR determines what counts as relevant evidence and what type of contradiction should be checked.

% This design gives \method{} its cross-domain invariance: the operator is reusable, but its visual meaning is domain-aware. For example, a temporal boundary always denotes a transition from inactive to active evidence, yet ``active'' corresponds to tool--tissue contact in surgery, object manipulation in industry, trajectory change in XSports, and social initiation in animal videos.

\subsection{Role-Decomposed Reasoning (RDR)}
Direct prompting asks the model to observe fine details, infer motion, compare options, and produce a valid label in a single generation. In EgoCross, this often causes early commitment: the model selects a plausible option before recording enough visual evidence. RDR reduces this failure mode by separating perception, dynamics, and decision into distinct reasoning roles when the sample requires temporal or behavior-heavy interpretation.

Given the normalized evidence $\mathcal{E}$ and the routed protocol $\mathcal{P}$, RDR produces intermediate textual states,
% After COR defines the evidence boundary, RDR prevents premature option commitment by separating observation from decision. Given $\mathcal{E}$ and $\mathcal{P}$, the backbone produces structured intermediate states,
\begin{align}
    z_p &= F_{\theta}(\mathcal{E},q,O;\mathcal{P}_{\mathrm{perc}}),\\
    z_d &= F_{\theta}(\mathcal{E},z_p,q,O;\mathcal{P}_{\mathrm{dyn}}),\\
    z_v &= F_{\theta}(\mathcal{E},z_p,z_d,q,O;\mathcal{P}_{\mathrm{ver}}),
    \label{eq:rdr}
\end{align}
where $F_{\theta}$ is the Qwen3-VL-4B backbone, $z_p$ records perception evidence, $z_d$ records temporal or behavioral dynamics, and $z_v$ records the option-oriented decision rationale. Motion-heavy cases benefit from this perception--dynamics--verification decomposition. For fine-grained small-object or controlled-vocabulary cases, \method{} uses a compact expert verifier $z_v=F_{\theta}(\mathcal{E},q,O;\mathcal{P}_{\mathrm{exp}})$ to avoid narrative drift. Thus, RDR adapts the reasoning depth while keeping the backbone fixed.

\begin{table*}[h]
\centering
\small
\setlength{\tabcolsep}{6pt}
\caption{Main results on EgoCross CloseQA. Baseline scores are copied from Table~2 of the EgoCross benchmark paper~\cite{egocross} and include all four domain-level accuracies. The challenge leaderboard provides the corresponding domain-level scores for \method{}.}
\vspace{-5pt}
\resizebox{0.7\linewidth}{!}{
\begin{tabular}{llccccc}
\toprule
Group & Method & Animal & XSports & Industry & Surgery & Overall\\
\midrule
\multirow{11}{*}{EgoCross CloseQA Baselines} & \multicolumn{6}{c}{\cellcolor{gray!8}\textit{Proprietary MLLMs}}\\
& GPT-4.1 & 64.48 & 43.09 & 45.71 & 57.24 & 52.63\\
& Gemini 2.5 Pro & 68.85 & 43.90 & 37.55 & 61.48 & 52.95\\
 & \multicolumn{6}{c}{\cellcolor{gray!8}\textit{Open-source MLLMs}}\\
& Qwen2.5-VL-3B & 41.53 & 36.59 & 36.33 & 35.69 & 37.54\\
& Qwen2.5-VL-7B & 53.55 & 41.87 & 37.55 & 46.29 & 44.82\\
& VideoLLaMA3-7B & 50.27 & 37.80 & 40.82 & 39.22 & 42.03\\
& InternVL3-8B & 49.18 & 41.06 & 33.06 & 47.00 & 42.58\\
 & \multicolumn{6}{c}{\cellcolor{gray!8}\textit{Egocentric MLLMs}}\\
& EgoVLPv2 & 24.04 & 23.17 & 34.69 & 26.50 & 27.10\\
& EgoGPT & 41.53 & 24.80 & 24.49 & 31.80 & 30.66\\
\midrule
\multirow{2}{*}{Challenge Submissions}
& \method{}, Source-Limited & \textbf{74.32} & \textbf{54.47} & \textbf{81.22} & \textbf{58.66} & \textbf{66.35}\\
& \method{}, Open-Source & \textbf{74.32} & \textbf{54.47} & \textbf{82.86} & \textbf{58.66} & \textbf{66.77}\\
\bottomrule
\end{tabular}
}
\vspace{-12pt}
\label{tab:main}
\end{table*}

\subsection{Boundary-aware Option Verification and Calibration}
Multiple-choice egocentric QA is vulnerable to plausible distractors. A candidate may share the correct object category but violate granularity, may occur near but not at the correct temporal boundary, or may be contradicted by an absence condition. BOV therefore treats each option as a hypothesis and asks whether it satisfies the evidence boundary defined by COR and supported by RDR. 

For an option $o_y$, the verification predicate is decomposed as
\begin{equation}
    \mathcal{V}(o_y)=
    \mathcal{S}(o_y)\wedge\mathcal{G}(o_y)\wedge\mathcal{T}(o_y)\wedge\neg\mathcal{K}(o_y),
    \label{eq:verification}
\end{equation}
where $\mathcal{S}$ denotes visual support, $\mathcal{G}$ denotes semantic-granularity consistency, $\mathcal{T}$ denotes temporal compatibility, and $\mathcal{K}$ denotes a hard contradiction. These terms are not computed by separate classifiers; they are the criteria enforced by the option-verification instruction. For example, a not-visible option must be absent throughout the evidence set, a temporal option must align with the inactive--active boundary, and a next-interaction option must agree with the final observed trajectory. The final answer is selected from the candidates that best satisfy Eq.~\eqref{eq:verification}; if multiple options remain plausible, the verifier favors the one with the strongest direct evidence and the fewest unsupported assumptions.

\subsection{Defensive Answer Calibration}
The final challenge is that a correct reasoning trace is not sufficient for a closed-set challenge submission: the answer must be recoverable as a valid option letter. DAC enforces this requirement by separating semantic decision from answer realization. The model is instructed to emit a constrained prediction field, and the final response is mapped into the option set through a hierarchical recovery rule,
\begin{equation}
    \hat{y}=\Gamma(r), \qquad \hat{y}\in\{A,B,C,D\},
    \label{eq:dac}
\end{equation}
where $r$ is the raw response and $\Gamma$ denotes deterministic recovery into the valid label space. In practice, valid structured outputs are accepted directly; otherwise, the system recovers the most reliable isolated option mention and falls back to a valid label only when no answer can be extracted. DAC does not alter visual semantics or add new evidence. Its role is to prevent avoidable leaderboard loss caused by formatting noise and to make the prediction interface stable.

Together, TEN, COR, RDR, BOV, and DAC form the \method. TEN answers \emph{when} and \emph{which frames} should be trusted; COR answers \emph{what capability} and \emph{which domain semantics} should be used; RDR answers \emph{what happened} in the evidence; BOV answers \emph{which option survives verification}; and DAC ensures that the verified decision is converted into a valid challenge submission.

%% file: 2-sec/3_exp.tex
\section{Experiments}
\subsection{Challenge Tracks and Baselines}
\method{} is evaluated in the Source-Limited and Open-Source tracks. Both tracks report accuracy for Animal, XSports, Industry, Surgery, and Overall. We compare with baselines from the EgoCross paper rather than other Codabench teams. The relevant baselines include proprietary MLLMs (GPT-4.1~\cite{gpt4}, Gemini 2.5 Pro~\cite{gemini15}), open-source MLLMs (Qwen2.5-VL-3B/7B~\cite{qwen25vl}, VideoLLaMA3-7B~\cite{videollama}, InternVL3-8B~\cite{internvl}), and egocentric-specific MLLMs (EgoVLPv2~\cite{egovlpv2}, EgoGPT~\cite{egogpt}). Since the challenge leaderboard is closed-ended, we report the CloseQA baseline results.

\begin{table*}[t]
\centering
\small
\vspace{-8pt}
\setlength{\tabcolsep}{5pt}
\renewcommand{\arraystretch}{1.15}
\caption{Ablation study of OmniEgo-R$^2$ showing cumulative performance gains over the 957 test questions. We start from a direct QA baseline and incrementally add our proposed modules. $\Delta$ Overall represents the absolute improvement compared to the Baseline.}
\vspace{-8pt}
\resizebox{0.74\linewidth}{!}{
\begin{tabular}{lcccccc}
\toprule
Model Variant & Animal & XSports & Industry & Surgery & Overall & $\Delta$ Overall \\
\midrule
Baseline (Direct MLLM) & 47.54 & 41.46 & 41.22 & 44.52 & 43.47 & -- \\
\quad + Temporal Evidence Normalization (TEN) & 71.58 & 46.75 & 45.71 & 53.00 & 53.08 & +9.61 \\
\quad + COR \& Semantic Bases & 71.58 & 55.69 & 56.73 & 52.30 & 57.99 & +14.52 \\
\quad + Role-Decomposed Reasoning (RDR) & 72.13 & 49.59 & 61.63 & 53.36 & 58.10 & +14.63 \\
\quad + BOV, DAC \& High-res (Full \method{}) & \textbf{74.32} & \textbf{54.47} & \textbf{82.86} & \textbf{58.66} & \textbf{66.77} & +23.30 \\
\bottomrule
\end{tabular}
}
\vspace{-16pt}
\label{tab:ablation_additive}
\end{table*}

\subsection{Implementation Details}
We use Qwen3-VL-4B checkpoints that are SFT separately for Animal, XSports, Industry, and Surgery. Frames are passed as timestamped image lists. The default visual budget uses up to $360K$ pixels, and small-object reasoning uses a higher budget in the industry setting. The maximum generation length is $2048$ tokens. Repetition penalty is set between $1.05$ and $1.1$.

% \section{Results}
\subsection{Main Results}
Table~\ref{tab:main} reports CloseQA baselines from the EgoCross paper~\cite{egocross} and the results of \method{}. \method{} ranks second in both challenge tracks and achieves the best overall accuracy among the compared methods. It improves over Gemini 2.5 Pro by $13.40$ and $13.82$ points in the source-limited and open-source results, and over Qwen2.5-VL-7B by $21.53$ and $21.95$ points. Since \method{} uses domain-SFT Qwen3-VL-4B checkpoints, the gain reflects both domain adaptation and structured test-time reasoning rather than a pure backbone comparison.

% \subsection{Domain-Level Analysis}
% The domain breakdown reveals different bottlenecks. Among official baselines, Animal Perspective is easiest: GPT-4.1 and Gemini 2.5 Pro reach $64.48$\% and $68.85$\%, and most open-source MLLMs also peak on Animal. In contrast, Industry and XSports are harder: the best baseline scores are only $45.71$\% and $43.90$\%, reflecting small components, controlled vocabularies, sparse frames, and fast viewpoint changes.

% \method{} changes this pattern substantially. Industry reaches $81.22$\%/$82.86$\%, more than $35$ points above the strongest Industry baseline, due to controlled vocabulary and small-object inspection. Animal also exceeds the strongest baseline, showing the benefit of self-other separation. XSports improves to $54.47$\%, but remains limited by high-speed temporal ambiguity. Surgery reaches $58.66$\%, above GPT-4.1 but below Gemini 2.5 Pro, suggesting that fine-grained tool ambiguity and occlusion still constrain the 4B backbone.

\subsection{Ablation Study}
To understand the source of these performance gains, we conduct a systematic ablation study. Instead of removing components, we start from a direct MLLM baseline and incrementally add our proposed modules to track the cumulative performance gains across the 957 challenge questions. Table~\ref{tab:ablation_additive} details these observations.

The most significant initial improvement comes from introducing Temporal Evidence Normalization (TEN). Compared to the direct MLLM baseline, adding TEN yields a substantial 9.61\% overall boost. Notably, the Animal domain accuracy jumps dramatically from 47.54\% to 71.58\%. This indicates that animal-perspective videos are especially sensitive to temporally grounded and reliability-aware observation, as crucial interaction cues often appear briefly or near the camera boundary. By converting sparse frames into timestamped, stable evidence units, TEN resolves the fundamental temporal ambiguity before any deep reasoning occurs.

Building upon TEN, adding the Capability-Oriented Router (COR) and Semantic Bases provides the next major gain, pushing the cumulative improvement to +14.52\% overall. This confirms our hypothesis regarding the structured invariance mechanism: normalizing heterogeneous domains into shared capabilities while preserving domain-specific visual grammar significantly improves cross-domain balance. Specifically, it raises Industry from 45.71\% to 56.73\% and XSports from 46.75\% to 55.69\%. 

Subsequently, integrating Role-Decomposed Reasoning (RDR) reveals an interesting domain trade-off. While the overall net gain is marginal (pushing the cumulative gain to +14.63\%), this obscures a significant internal shift. RDR acts as a ``slow-thinking'' mechanism that decoupling perception from decision-making. This greatly benefits complex procedural tasks, pushing Industry accuracy up to 61.63\%. However, it comes at the cost of XSports performance, which drops from 55.69\% to 49.59\%, suggesting that explicit role decomposition may overly strict for rapid, motion-heavy athletic scenarios that rely on continuous trajectory contexts rather than discrete semantic states.

Finally, incorporating Boundary-aware Option Verification (BOV), Defensive Answer Calibration (DAC), and high-resolution small-object mode (forming our full \method{}) resolves these bottlenecks and delivers a massive final surge, bringing the total cumulative gain to +23.30\%. The most striking improvement is seen in the Industry domain, which skyrockets to 82.86\%, and XSports recovers to 54.47\%. This demonstrates that full option-wise hypothesis checking (BOV) and adaptive resolution are critical for distinguishing subtle semantic granularity mismatches among distractors. Concurrently, DAC prevents avoidable leaderboard losses from malformed or unstable closed-set outputs. Overall, these modules are highly complementary, and their combined integration enables \method{} to achieve its optimal performance of 66.77\%.

%% file: 2-sec/4_conclusion.tex
\section{Conclusion}
This report presented \method{} for the 1st Cross-Domain EgoCross Challenge at CVPR 2026. The method addresses temporal boundary ambiguity, cross-domain semantic granularity mismatch, and decision instability under close options through temporal evidence normalization, capability routing, structured reasoning, option verification, and defensive calibration. \method{} achieves second place in both the Source-Limited and Open-Source tracks, with 66.35\% and 66.77\% overall accuracy, respectively. The results and planned ablations suggest that careful test-time reasoning design can substantially improve MLLM robustness on cross-domain egocentric video QA.